\documentclass{article}

\RequirePackage{fix-cm}

\usepackage{arxiv}

\usepackage[utf8]{inputenc} 
\usepackage[T1]{fontenc}    
\usepackage{hyperref}       
\usepackage{url}            
\usepackage{booktabs}       
\usepackage{amsfonts}       
\usepackage{nicefrac}       
\usepackage{microtype}      
\usepackage{lipsum}
\usepackage{subfig}
\usepackage{amsmath}
\usepackage{natbib}
\usepackage{lineno}
\usepackage[dvipsnames]{xcolor}
\usepackage{mathptmx}      
\usepackage{latexsym}
\usepackage{array,multirow,graphicx}

\title{Prediction of Porosity and Permeability Alteration based on Machine Learning Algorithms}

\author{
  Andrei Erofeev \\
  Skolkovo Institute of Science and Technology\\
  Skolkovo Innovation Center\\
  Building 3, Moscow, 143026, Russia\\
  \texttt{Andrei.Erofeev@skoltech.ru}
   \And
 Denis Orlov \\
  Skolkovo Institute of Science and Technology\\
  Skolkovo Innovation Center\\
  Building 3, Moscow, 143026, Russia\\
  \texttt{D.Orlov@skoltech.ru}
  \And
 Alexey Ryzhov \\
  Gazprom VNIIGAZ LLC\\
  box 130, Moscow, 115583, Russia\\
  \texttt{A\textunderscore Ryzhov@vniigaz.gazprom.ru}
  \And
 Dmitry Koroteev \\
  Skolkovo Institute of Science and Technology\\
  Skolkovo Innovation Center\\
  Building 3, Moscow, 143026, Russia\\
  \texttt{D.Koroteev@skoltech.ru}
}

\begin{document}
\maketitle

\begin{abstract}
The objective of this work is to study the applicability of various Machine Learning algorithms for prediction of some rock properties which geoscientists usually define due to special lab analysis. We demonstrate that these special properties can be predicted only basing on routine core analysis (RCA) data. 
To validate the approach core samples from the reservoir with soluble rock matrix components (salts) were tested within 100+ laboratory experiments. The challenge of the experiments was to characterize the rate of salts in cores and alteration of porosity and permeability after reservoir desalination due to drilling mud or water injection. For these three measured characteristics, we developed the relevant predictive models, which were based on the results of RCA and data on coring depth and top and bottom depths of productive horizons. 
To select the most accurate Machine Learning algorithm a comparative analysis has been performed. It was shown that different algorithms work better in different models. However, two hidden layers Neural network has demonstrated the best predictive ability and generalizability for all three rock characteristics jointly. The other algorithms, such as Support Vector Machine and Linear Regression, also worked well on the dataset, but in particular cases.
Overall, the applied approach allows predicting the alteration of porosity and permeability during desalination in porous rocks and also evaluating salt concentration without direct measurements in a laboratory. 
This work also shows that developed approaches could be applied for prediction of other rock properties (residual brine and oil saturations, relative permeability, capillary pressure, and others), which laboratory measurements are time-consuming and expensive.

\keywords{machine learning \and routine and special core analysis \and reservoir properties \and salted formations \and porosity and permeability alteration  }
\end{abstract}

\section{Introduction}
\label{intro}
Laboratory study of reservoir rock samples of a geologic formation (Core Analysis) is the direct way to determine reservoir properties and to provide accurate input data for geological models \citep{Andersen2013CoreTruth}. Geoscientists have developed a variety of approaches for measuring properties of reservoir rocks, such as porosity, permeability, residual oil saturation, and many others. The information obtained from core analysis aids in formation evaluation, reservoir development, and reservoir engineering \citep{mcphee2015core,mahzari2018co}. It can be used to calibrate log and seismic measurements and to help with well placement, completion design, and other aspects of reservoir production. 

Common applications of Core Analysis include \citep{Gaafar2015}:
\begin{itemize}
\item definitions of porosity and permeability, residual fluid saturations, lithology and prediction of possible production of gas, condensate, oil or water; 
\item definition of spatial distributions of porosity, permeability and lithology to characterize a reservoir in macro scale;
\item definition of fluids distribution in a reservoir (estimation of fluids contacts, transition zones); 
\item performing special core analysis tests to define the most effective field development plan to maximize oil recovery and profitability. 
\end{itemize}

Unfortunately, Core Analysis is expensive and tedious. Laboratory study requires careful planning to obtain data with minimum uncertainties \citep{BardOttesen2008Core}. Proper results of basic laboratory tests, provides the reservoir management team with a vital information for further development and production strategy. 

Core analysis is generally categorized into two groups: conventional or routine core analysis (RCA) and special core analysis (SCAL) \citep{dandekar2006petroleum}. RCA generally refers to the measurements for defining porosity, grain density, absolute permeability, fluid saturations, and a lithologic description of the core. Samples for conventional core analysis are usually collected three to four times per meter \citep{monicard1980properties}. Fine stratification features and spatial variations in lithology may require more frequent sampling. 

Probably the most prominent SCAL tests are two-phase or three-phase fluid flow  experiments in the rock samples for defining relative permeability, wettability, and capillary pressure. In addition, SCAL tests may also include measurements of electrical and mechanical properties, petrographic studies and formation damage tests \citep{orlov2018self}. Petrographic and mineralogical studies include imaging of the formation rock samples through thin-section analysis, X-ray diffraction, scanning electron microscopy (SEM), and computed tomography (CT) scanning in order to obtain better visualization of the pore space \citep{dandekar2006petroleum,liu2017pore,soulaine2016micro}. SCAL is a detailed study of rock characteristics, but it is time-consuming and expensive. As a result, a number of SCAL measurements is much less than a number of RCA measurements (5-30\% of RCA tests). In this way, SCAL data space requires correct expansion or extrapolation  to the data space covered by RCA. To provide the expansion, core samples set used in SCAL tests should be highly representative and contain all the rock types and cover a wide range of permeability and porosity \citep{stewart2011well}. Even then, sometimes it is difficult to estimate correlations between conventional and special core analysis results and expand SCAL data to the available RCA dataset. There are few common approaches on stretching the SCAL data to RCA data space:
\begin{itemize}
\item typification (defining rock types with typical SCAL characteristics in certain ranges of RCA parameters);
\item petrophysical models (SCAL characteristics included as parameters in functional dependencies between RCA characteristics);
\item prediction models based on machine learning (RCA parameters used as features to predict SCAL characteristics).
\end{itemize}

The first approach leads to a significant simplification of reservoir characterization and is based on subjective conclusions. Petrophysical models allow predicting only a few of SCAL characteristics (basically capillary curves and residual saturations). The last approach looks more promising as it accounts for all the available features (measurements) and builds implicit correlations among the features \citep{meshalkin2018robotized,tahmasebi2018rapid}.

The purpose of this research is to demonstrate the performance of Machine Learning (ML) at maximizing the effect of RCA and SCAL data treatment. Machine Learning is a subarea of artificial intelligence based on the idea that an intelligent algorithm can learn from data, identify patterns and make decisions with minimal human intervention \citep{kotsiantis2007supervised}.

Commonly spread feature of fields in Eastern Siberia is salts (the ionic compound that can be formed by the neutralization reaction of an acid and a base) presented in the pore volume of the deposits. Salts distribution in the reservoirs depends on a complex of sedimentation processes. Thus, the key challenge of this work is to develop prediction models, which can characterize the quantity of soluble rock matrix components (sodium chloride and other ionic compounds) and an increase of porosity and permeability after reservoir desalination due to drilling mud or water injection (ablation). 

One of the main challenges for geoscientists is forecasting salts distribution in productive horizons together with porosity and permeability alteration due to the salts ablation. It is very important for:

\begin{itemize}
\item estimations of original porosity and permeability in wells as water-based drilling muds can change pore structure during wellbore drilling and coring,
\item RCA and SCAL data validation and correction due to pore structure alteration during core sample preparation and consequent measurements,
\item reservoir engineering (IOR\&EOR based on water injection).
\end{itemize} 

In this work salts content and alteration of porosity and permeability after desalination could be considered as a SCAL measurements because it is expensive and time-consuming. The procedure of alteration estimation  includes porosity and permeability measurements before and after water injection in core samples and its desalinization during long-term one phase water filtration. Porosity and permeability before desalination, sample density and lithology and texture description are the RCA input data for our predictive models.

The significant benefit of ML predictive models is that one may not have to perform SCAL measurements for all the core samples, but can conduct prediction of the results \citep{unsal2005genetic}. Once a predictive model of any SCAL results is trained it could be effectively used for future forecasting. There are a lot of ML algorithms to build a predictive model \citep{hastie2001elements}. In our work we used the following algorithms: linear regression (with and without regularization) \citep{boyd2004convex, freedman2009statistical}, decision tree \citep{Quinlan1986}, random forest \citep{ho1995random}, gradient boosting \citep{Friedman2000Greedy}, neural network \citep{haykin1994neural} and support vector machines \citep{Cortes1995SVN}. The choice of algorithm strongly depends on the considered problem, data quality and size of the dataset. For example, it would be unnecessary to build convolutional or recursive NN in our problem due to the small dataset size and its structure. However, more simple algorithms (mentioned above) could be adopted for discussed cases.

Accordingly, we have two goals in our research. First is to develop a predictive model of salts concentration using information of RCA and some additional data about coring depth and top and bottom depths of productive horizons. Second is to develop relevant predictive models of porosity and permeability.

The main innovation elements of the research are:
\begin{itemize}
\item Special experimental investigations of porosity and permeability increasing in core waterflooding tests,
\item Validation of predictive algorithms to define the best predictive model,
\item Accounting 10 features of core samples to predict porosity and permeability after rock desalination and 9 features to predict salts content;
\item The high quality of models for prediction of porosity and permeability after rock desalination and rather good quality of model for evaluation of rock salinity.
\end{itemize}

\section{Materials and Methods}
\label{sec:2}

\subsection{Hydrocarbon reservoir characterization}
\label{sec:2.1}

The Chayandinskoye oil and gas condensate field is located in the Lensk district of Sakha (Yakutia) Republic in Russia and hosted towards the south of the Siberian platform within the Nepa arch. The field belongs to the Nepa-Botuobinsky oil and gas area, which contains rich hydrocarbons reserves. The main gas and oil resources are associated with the Vendian terrigenous deposits (Talakh, Khamakin and Botuobinsk horizons) which are overlapped by a thick series of the salt-bearing sediments. 

Chayandinskoye field is characterized by a complex geological structure and special thermobaric formation conditions (reservoir pressure of 36-38 MPa, overburden stress of 50 MPa, temperature of 11-17$^{\circ}$C). The Vendian deposits consist predominantly of quartz sandstones and aleurolits with a low level of cementation and development of indentation and incorporation of grains. Another essential feature of the field is salts presented in the pore volume of the deposits. Salts distribution in the reservoir is exceptionally irregular due to various sedimentation processes: change in thermobaric condition during regional uplifts and erosional destruction of deposits, paleoclimate cooling and glaciation, in addition to filtration of brines through rock faults and fractured zones \citep{Ryzhov2014filtration}.  Usually, the most common salt in rock matrix is sodium chloride (NaCl), but many other salts occur in varying smaller quantities. The same conclusion based on TDS analysis (measurement of the total ionic concentration of dissolved minerals in water) is correct for brine composition.
Rocks analysis demonstrates that highly salinized formations are coarse-grain poorly sorted rocks with mass salts concentration – ranging from 4 to 30\%. The porosity of the salted rocks is 1 - 8\% (seldom $\geq$ 10\%). After core desalination permeability could be increased up to 60 times and porosity - up to 2.5 times.

\subsection{Dataset}
\label{sec:2.2}
We included the following features to the dataset for our prediction models:

\begin{itemize}
\item measurements of salts mass concentrations for core samples with various values of initial porosity and permeability, lithology, depth, horizons' ID and wells' ID;
\item measurements of porosity and absolute permeability before and after desalination.
\end{itemize}

All tests are performed on 102 cylindrical core samples with 30 mm radius and 30 mm length. Sample preparation included delicate extraction in the alcohol-benzene mixture at room temperature (to avoid premature desalination) and drying up to constant weight. Absolute permeability was measured at ambient conditions in the steady-state regime of nitrogen flow. Porosity was determined by a gas-volumetric method based on Boyle’s Law \citep{rp401998recommended}.

\begin{figure*}[h!]

\begin{minipage}{.5\linewidth}
\subfloat[]{\label{fig1:a}\includegraphics[scale=.22]{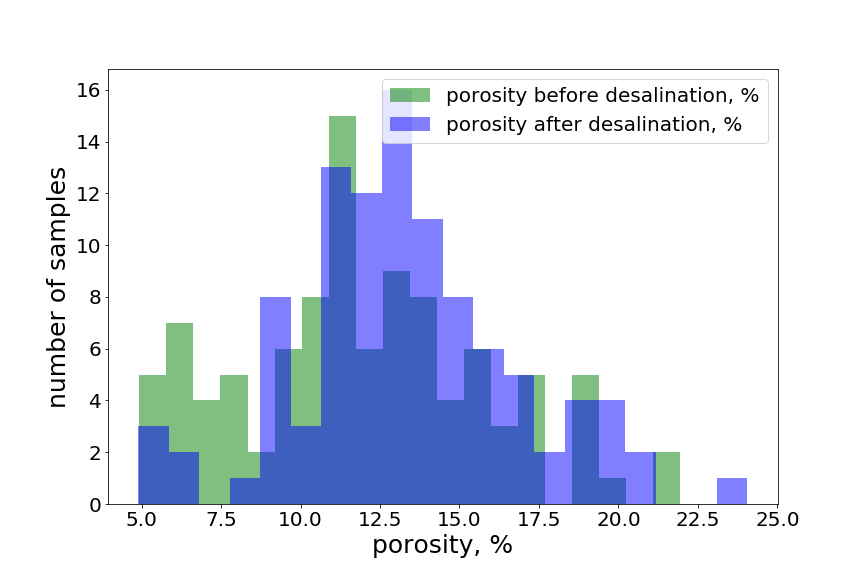}}
\end{minipage}%
\begin{minipage}{.5\linewidth}
\subfloat[]{\label{fig1:b}\includegraphics[scale=.22]{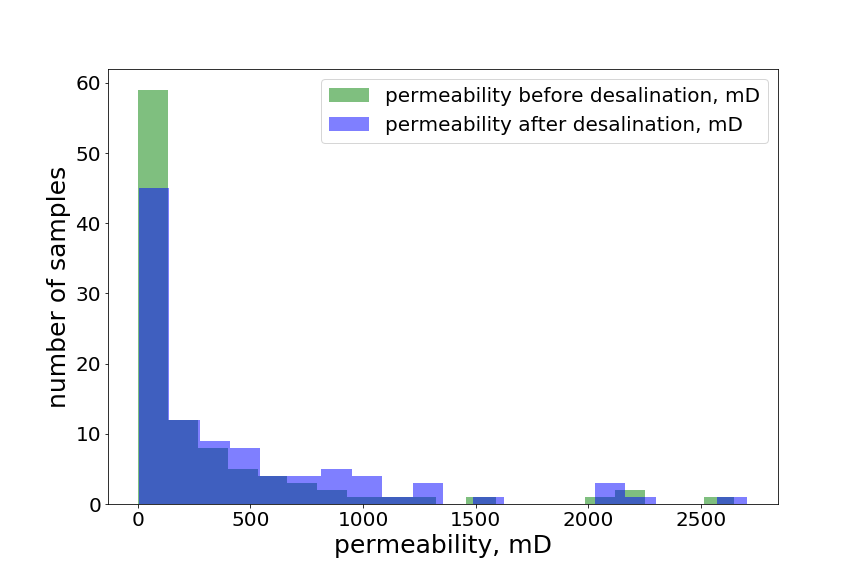}}
\end{minipage}\par\medskip
\subfloat[]{\label{fig1:c}\includegraphics[scale=.22]{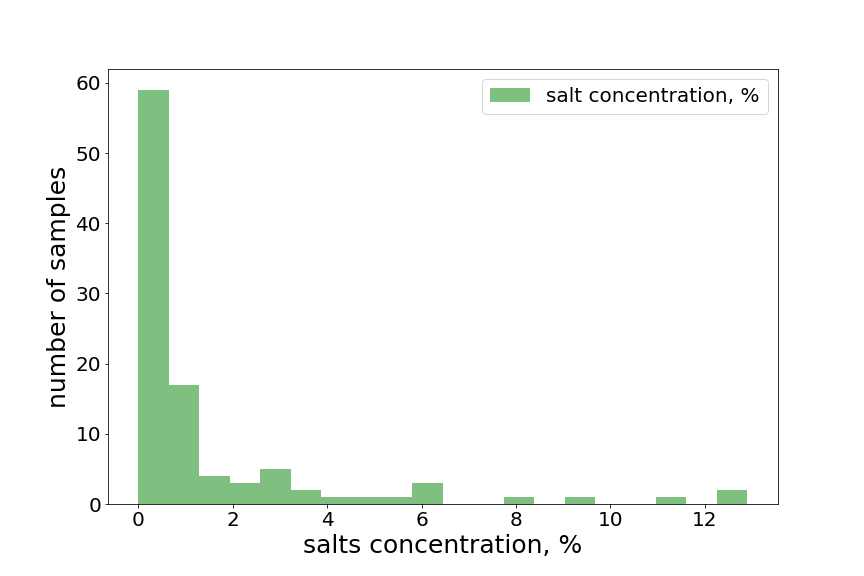}}
\centering
\caption{Data set. (a) – permeability distribution, (b) – porosity distribution, (c) – salts concentration distribution.}
\label{fig1}
\end{figure*}

For salts ablation, we injected in each core sample more than 10 pore volumes of brine with low salinity (30 mg/cc). To enhance ablation, we also performed additional extraction in the alcohol-benzene mixtureto remove oil films preventing salts dissolution. After that, the samples were dried up to a constant weight to measure porosity and permeability after desalination. Results of porosity and permeability measurements (before and after salts ablation) are presented in figures ~\ref{fig1:a}, ~\ref{fig1:b}.

Measurements of salts mass concentrations for core samples were based on the data of sample weighting before and after desalination. The resulting expression for salts concentration could be defined as:
\begin{equation}\label{eq1:saltC}
C_{salt} = \frac{\Delta m}{m_0},
\end{equation}
where $\Delta m$ - is the sample mass difference before and after ablation (g); $m_0$ - sample mass before desalination (g). All the measurements were made on oven-dry samples. Results of salts mass concentrations measurements presented in figure ~\ref{fig1:c}.

For salts concentration predictive model, we have used 9 features: formation top depth, formation bottom depth, initial (before desalination) porosity and permeability, sample depth adjusted to log depth, sample density (before desalination), average grain size (by lithology and texture description), sample colour and horizon ID. Average grain size was quantified from textual lithology description in the following way:

Gravel –- 1 mm;

Coarse sand -– 0.5 mm;

Medium sand –- 0.25 mm;

Fine sand -– 0.1 mm;

Coarse silt –- 0.05 mm;

Fine silt –- 0.01 mm;

Clay –- 0.005 mm.

For sample colour and horizon type, we have used the classification scheme containing 6 colour types and 3 horizon types. If the sample has any of 6 colours and any of 3 horizons we mark “1”, otherwise, we mark “0”.

For porosity and permeability predictive models we have used 9 previously described features plus salts concentration. All 10 features accounting in machine learning algorithms are presented in table~\ref{tab:1}. 

\begin{table}[ht!]
\caption{All features used in predictive models}
\label{tab:1}       
\begin{tabular}{lll}
\hline\noalign{\smallskip}
No. & Feature & Unit  \\ \hline
    1 & salts concentration & g/g \\ \hline
	2 & formation top depth & m \\ \hline
    3 & formation bottom depth & m \\ \hline
    4 & porosity before desalination & \% \\ \hline
    5 & absolute permeability before desalination & mD \\ \hline
    6 & sample depth & m \\ \hline
    7 & sample density & g/cc \\ \hline
    8 & average grain size & mm \\ \hline
    9 & color & -* \\ \hline
    10 & depth horizon & -* \\ \hline
     \multicolumn{3}{l}{$*$ dimensionless}
\end{tabular}
\end{table}

\subsection{Prediction models}
\label{sec:2.3}
We have used 9 models: linear regression (simple, with L1 and L2 regularization), decision tree, random forest, gradient boosting (two different implementations with and without regularization) and neural network, support vector machines to compare their predictive power.

\textbf{Linear regression} \citep{hastie2001elements,freedman2009statistical}. Simple linear regression expresses predicting value as linear combination of the features:
\begin{equation}\label{eq2:LR}
y = w_0x_0 + w_1x_1 + ... + w_px_p + b
\end{equation}
where $y$ is predicting parameter; $x$ is a vector of features; $\mathbf{w}$ is a vector of optimizing coefficients. 

The optimization problem for regression is given by the expression:
\begin{equation}\label{eq3:OP_LR}
\min_{\textbf{w}, b} F(\textbf{w}, b) = \frac{1}{m} \sum_{i=1}^{m} (\textbf{wx} + b - y_i)^2
\end{equation}
Coefficients $\mathbf{w}$, $b$ are defined from a training set of data ($y_i$ is the actual value of the predicting parameter; $m$ is the size of the training set).  Further, the linear regression model with predefined coefficients could be effectively applied to fit new data. This algorithm is implemented in LinearRegression() method of Python scikit-learn library \citep{Pedregosa2012Scikit-learn}. 

Sometimes regression with regularization works better than simple regression. In a case when we have many features, linear regression procedure leads to overfitting: enormous weights $\textbf{w}$ that fit the training data very well, but poorly predicts future data.  “Regularization” means modifying the optimization problem to prefer small weights. To avoid the numerical instability of the Least Squares procedure regression with L2 and L1 regularizations are often applied \citep{hastie2001elements}.

\textbf{Linear regression with L2 regularization (Ridge)} \citep{boyd2004convex}. This approach is based on Tikhonov regularization, which addresses the numerical instability of the matrix inversion and subsequently produces lower variance models:
\begin{equation}\label{eq4:LR_L2}
\min_{\textbf{w}, b} F(\textbf{w}, b) = \frac{1}{m} \sum_{i=1}^{m} (\textbf{wx} + b - y_i)^2 + \lambda ||\textbf{w}||^2_2
\end{equation}
All variables have the same meanings as in linear regression case. Optimal regularization parameter $\lambda$ is chosen in a way to get the best model fitting while weights $\textbf{w}$ are small. This algorithm is implemented in Ridge() method of Python scikit-learn library \citep{Pedregosa2012Scikit-learn}. 

\textbf{Linear regression with L1 regularization (Lasso)} \citep{boyd2004convex}. While L2 regularization is an effective approach of achieving numerical stability and increasing predictive performance, it does not address another problem with Least Squares estimates, parsimony of the model and interpretability of the coefficient values \citep{Tibshirani2011Regression}. Another trend has been to replace the L2-norm with an L1-norm:
\begin{equation}\label{eq5:LR_L1}
\min_{\textbf{w}, b} F(\textbf{w}, b) = \frac{1}{m} \sum_{i=1}^{m} (\textbf{wx} + b - y_i)^2 + \lambda ||\textbf{w}||_1
\end{equation}

This L1 regularization has many of the beneficial properties of L2 regularization, but yields sparse models that are more easily interpreted \citep{hastie2001elements}. L1 regularization algorithm is implemented in Lasso() method of Python scikit-learn library \citep{Pedregosa2012Scikit-learn}.

\begin{figure}[ht!]
\begin{minipage}{.5\linewidth}
\subfloat[]{\label{fig2:a}\includegraphics[scale=.18]{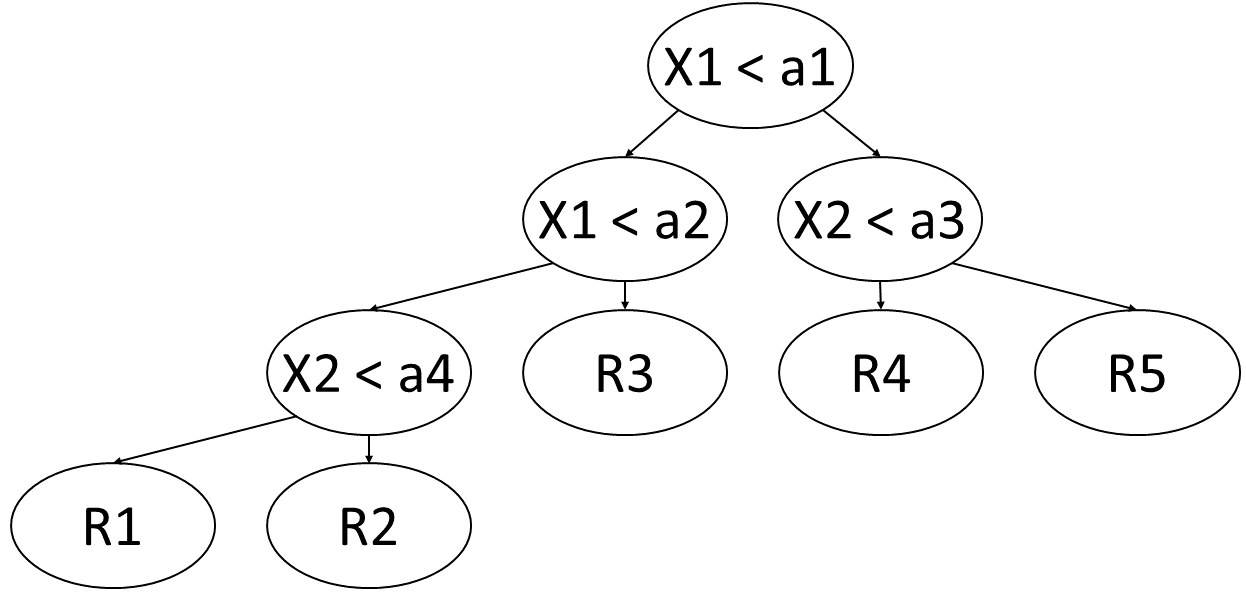}}
\end{minipage}%
\begin{minipage}{.5\linewidth}
\subfloat[]{\label{fig2:b}\includegraphics[scale=.18]{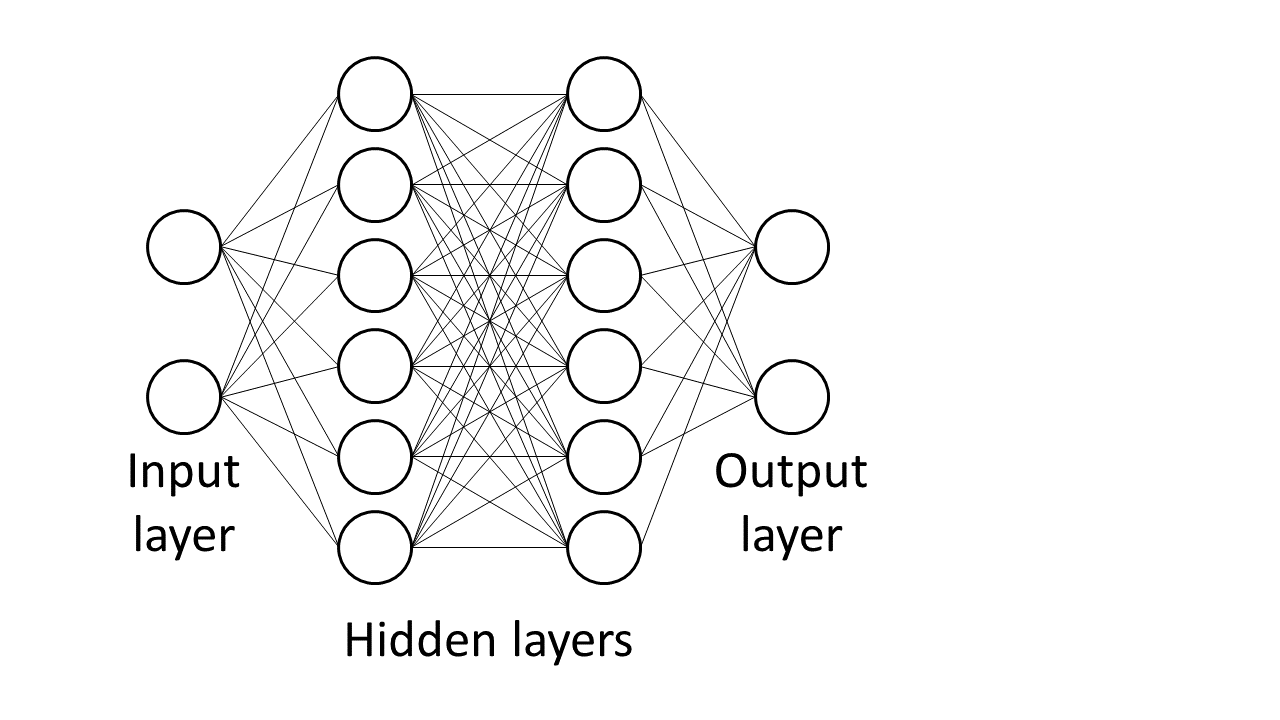}}
\end{minipage}\par\medskip
\caption{Examples of prediction algorithms. (a) – Decision tree example, (b) – Scheme of artificial neural network with two hidden layers}
\label{fig2}
\end{figure}

\textbf{Decision Tree} \citep{Quinlan1986}. Decision tree is a tree representation of a partition of feature space. There are numbers of different types of tree algorithms, but here we will consider only CART (Classification and Regression Trees) \citep{breiman2017classification} approach. A classification tree is a decision tree which returns a categorical answer (class, text, color and other) while a regression tree is a  decision tree which responses with an exact number. Figure ~\ref{fig2:a} demonstrates a simple example of the decision tree in a case of two-dimensional space based on two features X1 and X2.

A decision tree consists of sets of leaves and nodes. One may builts very detailed deep tree with many nodes however such tree will suffer from overfitting. Usually, the maximum depth of the tree is restricted. To build a tree, the recursive partition is applied until a sufficient size of a tree would be obtained. The criteria for tree splitting is often given by Gini index or information gain criteria in classification case and mean squared error or mean absolute error in the regression case. These functions are used to measure the quality of a split and choose the optimal point of partition. Tree construction could involve not all input variables, so tree managed to demonstrate which variables are relatively important, but it could not rank the input variables.
Decision tree algorithm exploited in this work is implemented in DecisionTreeRegressor() method of Python scikit-learn library \citep{Pedregosa2012Scikit-learn}.

\textbf{Random Forest} \citep{ho1995random}, \citep{Breiman2001}. The main idea of this method is to build many independent decision trees (ensemble of trees), train them on data subset and receive predictions. The algorithm uses bootstrap re-sampling to prevent overfitting. Bootstrapping is a re-sampling with replacement: bootstrap sets are built on initial data, where several samples are replaced with other repeating samples. Each tree is built on individual bootstrap set (so, for N tree estimators, we need N different bootstrap representations). Consequently, all trees are different as they are built on different datasets and hold different predictions.  Then all trees are aggregated together after training and the final prediction is obtained by averaging (in the case of regression) predictions of each tree. One useful feature of Random Forest algorithms is that it could rank input features.  It is implemented in RandomForestRegressor() method of Python scikit-learn library \citep{Pedregosa2012Scikit-learn}.  

\textbf{Gradient Boosting} \citep{Friedman2000Greedy}. This method uses "boosting" of the ensemble of weak learners (often decision trees). Boosting algorithm combines trees sequentially in such a way that the next estimator (tree) learns from the error of previous one: this method is iterative, and each next tree is built as a regression on pseudo-remainders. Similar to any other ML algorithm, Gradient Boosting uses loss function to minimize. Also, gradient descent is applied to minimize error (loss function) associated with adding a new estimator. The final model is obtained by combining the initial estimation with all subsequent estimations with appropriate weights. Gradient boosting method used in this study implemented in the scikit-learn library in GradientBoostingRegressor() method \citep{Pedregosa2012Scikit-learn}. Also, XGBoost library \citep{chen2016xgboost} was considered because it allows adding regularization to the model.

\textbf{Support Vector Machine (SVM)} \citep{Cortes1995SVN}. The idea of Support vector machine (in case of regression) is to find a function that approximates data in the best possible way. This function has at most $\epsilon$ deviation out from real train values $y_i$ and as flat as possible. Such linear function $f(\textbf{X})$ could be expressed as:

\begin{equation}\label{eq11:SVM_linfun}
f(\textbf{X}) = w\textbf{X} + b
\end{equation}

The optimization problem could be formulated in the following form:

\begin{equation}\label{eq12:SVM_OP}
\begin{split}
&\min_{W} \frac{1}{2}||w||^2 \\
subject & \: to: 
\begin{cases}
  y_i - w \textbf{X} - b \leq \epsilon\\    
  w \textbf{X} - b - y_i \leq \epsilon
\end{cases}
\end{split}
\end{equation}

Often, some errors beyond $\epsilon$ are allowed, which requires introducing of slack variables $\xi_i, \xi_i^*$ into the problem:
\begin{equation}\label{eq12b:SVM_OP}
\begin{split}
&\min_{W} \frac{1}{2}||w||^2 + C\sum_{i = 1}^{m}(\xi_i + \xi_i^*) \\
subject & \: to: 
\begin{cases}
  y_i - w \textbf{X} - b \leq \epsilon + \xi_i\\    
  w \textbf{X} - b - y_i \leq \epsilon + \xi_i^*\\
  \xi_i, \xi_i^* \geq 0
\end{cases}
\end{split}
\end{equation}

The solution of SVR problem is usually given in a dual form\citep{hsieh2008dual}, which includes calculation of Lagrange multipliers $\alpha_i, \alpha'_i$. In this formulation solution looks as follows: 

\begin{equation}\label{eq13:SVM_DP}
f(x) = \sum_{i = 1}^{m}(\alpha'_i - \alpha_i)K(x_i, x) + b
\end{equation}

where $K(x_i, x)$ is a kernel \citep{crammer2001algorithmic}.

In this work, the standard Gaussian kernel has been applied:

\begin{equation}\label{eq15:SVM_GaussKernel}
K(x, x') = exp\left(-\frac{||x - x'||^2}{2 \sigma ^ 2}\right)
\end{equation}

SVM algorithm is implemented in SVR() method of Python scikit-learn library \citep{Pedregosa2012Scikit-learn}.

\textbf{Neural Network} \citep{haykin1994neural, Schmidhuber2014Deep}. Artificial neural network (ANN) is a mathematical representation of biological neural network (figure ~\ref{fig2:b}). It consists of several layers with units that are connected by links \citep{mcculloch1943logical}. Each link has associated weight and activation level. Each node has an input value, an activation function and an output. In ANN information propagates (forward pass) from first (not hidden) layer with inputs to next hidden layer and then to further hidden layers until the output layer would be achieved. The value in each node of the first hidden layer obtained after calculation of activation function for the dot product of inputs and weights. Next hidden layer receives the output of the previous one and puts its dot product with weights to the activation an so on. Initially, all weights for all nodes are assigned randomly. ANN calculates first output with random weights. Then compare it to real value, calculate the error and adopts weights to obtain smaller error on the next iteration via backpropagation (ANN training algorithm). After training all weights are tuned, and one may make a prediction for a new data by passing them into an ANN which will calculate output via forward propagation throw all activations. \citep{rumelhart1986learning}. Scikit-learn library provides ANN representation in MLPRegressor() method \citep{Pedregosa2012Scikit-learn}. This implementation allows to indicate the number of hidden layers, number of nodes in each layer, activation function, learning rate and some other parameters.

\subsection{Methodology of using machine learning algorithms}
\label{sec:2.4}
\textbf{Metrics.} To evaluate the accuracy of applied methods and compare them between each other, the following metrics have been exploited.

The coefficient of determination R2 is the proportion of the total (corrected) sum of squares of the dependent variable “explained” by the independent variables in the model. R2 score is a part of dispersion of dependent variable that is predictable by independent variables: 

\begin{equation}\label{eq16:R2}
R^2 = 1 - \frac{\sum_{i}(y_i - \hat y_i)^2}{\sum_{i}(y_i - \bar y)^2}\\
\end{equation}

where $y_i$ are real values, $\hat y_i$ are predicted values, $\bar y$ is a mean value.

Mean squared error corresponds to quadratic errors:

\begin{equation}\label{eq17:MSE}
MSE(y, \hat y) = \frac{1}{n_{samples}} \sum_{i = 0}^{n_{samples} - 1} (y_i - \hat y_i)^2
\end{equation}

Mean absolute error corresponds to absolute errors:
\begin{equation}\label{eq18:MAE}
MAE(y, \hat y) = \frac{1}{n_{samples}} \sum_{i = 0}^{n_{samples} - 1} |y_i - \hat y_i|
\end{equation}

\textbf{Cross-validation.} Since this study is limited in the amount of data, cross-validation (CV) has been applied. There are several cross-validation techniques. For example, k-fold, when we split whole data into k parts, then use first part for testing the performance of ML model after training it on the other k-1 parts. Next, we could take the second part for testing and the rest parts for training and so on k times. In the end we will have k different values of metrics - R2, MAE, MSE (eqs. (\ref{eq16:R2}) –- (\ref{eq18:MAE})) calculated for each fold. So, via cross-validation, we could obtain mean values of metrics and their standard deviation. Another cross-validation technique called random permutation supposes random division of data into train and test set, then data shuffle and we can obtain new division on test and train set. This process repeats n times and each time metrics are calculated.
Similarly, in the end, we can evaluate the mean values of metrics and standard deviations. So, cross-validation allows not only calculate R2 (or MAE and MSE) for the test set but do it several times by independent data splitting into test and train sets. Since in our task we are restricted in the amount of data, cross-validation was used several times (for hyperparameters tuning, model's performance estimation and making predictions to plot predicted results versus real data). Models building and estimation was done in 3 steps:
\begin{enumerate}
    \item Hyperparameters tuning was done with the help of exhaustive grid search \citep{bergstra2012random}. This process allows to search through given ranges of each hyperparameter and define optimal values which led to the best R2 (or MAE, MSE etc.) scores. This process implemented in scikit-learn Python library \citep{Pedregosa2012Scikit-learn} in method GridSearchCV(). The method simply calculates CV score for each combination of hyperparameters in a given range. Random permutation approach with 10 repeats was chosen as CV iterator. GridSearchCV() allows not only find the best hyperparameters but also calculate metrics in the optimal point. However, just 10 repeats are not enough for very accurate final estimation of mean value and deviation, while taking more repeats require more computational time. So, final estimation with known hyperparameters would be done next. 
    \item Evaluation of the ML model with optimal hyperparameters (defined on the first step) was also done via CV with random permutation approach. However, here, we take 100 repeats, and it is enough for a fair result.
    \item To plot predictions versus real values, we applied k-fold CV. In our particular case, we took k equals the number of samples. So, first of all, we train our model on all data without one point, then predict at this point (testing) by trained model. Next, we take the second point - remove it from the dataset and train model from scratch, then obtain prediction in point. This process was made for all data points (102 times). It allows to obtain predictions for all points and compare them to initial data visually. 
\end{enumerate}

Some researchers \citep{choubineh2019estimation} suggest the other way to validate the quality of ML models. The data records of the dataset are divided into training, validation and testing subsets, respectively. Where validation set is used for hyperparametres tuning while training on training set and test set is used for final model evaluation.   However, unfortunately, a single random partition of the data on subsets could not be enough for correct model estimation due to non-uniformity of the dataset. Another random partition will give another value of metrics (R2, MAE, MSE and others). Single partition of the data is reasonable only in case of the big size of the dataset.

\textbf{Normalization of data.} Some machine learning methods require normalized data to proceed correctly (SVM and Neural Network), so in these cases, the data has been normalized by using the mean and standard deviation of the training set:

\begin{equation}\label{eq19:normalize}
X_{scaled_i} = \frac{X_i - \bar X}{\sigma_X}
\end{equation}
where $X_i$ is the feature vector, $\bar X$ is the mean value of feature vector, $\sigma_X$ is a variance.

\section{Results \& Discussion}
\label{sec:3}

In this section results of porosity, permeability and salts concentration predictions are presented, analyzed and compared. We made predictions of these reservoir properties on the basis of features described in section 2.2 by algorithms described in section 2.3. However, only Linear regression with L1 regularization was taken into account out of 3 Linear regression algorithms (because algorithms are very similar, implemented within the same library and results are very close, and regression with L1 regularization performs slightly better in our task). Also, two different libraries for Gradient boosting calculation was applied (they reported separately) because XGBoost library allows regularization while scikit-learn not. So, we reported results for 7 different algorithms. Each algorithm was adopted to the experimental data to get the best performance due to the cross-validation procedure. 

Since our dataset is not such big from Big Data point of view and contains different measurements errors, it turns out that the proportion of its splitting via cross-validation procedure is important. We found out via grid-search that for our small dataset would be optimal to left 35\% of data for testing on each cross-validation pass to obtain the estimation of algorithm performance.

\subsection{Porosity prediction}
\label{sec:3.1}

Porosity model was the first ML model we built. Actually, the material balance equation defines specific dependence between porosity and salts concentration:

\begin{equation}\label{eq20:salt_conc}
\phi = \phi_0 + \frac{\rho_{r}^{0}}{\rho_{ha}}C_{salt}
\end{equation}

where $\phi$ is porosity after desalinization; $\phi_0$ is porosity before desalinization; $C_{salt}$ is the mass salts concentration;  $\rho_{ha}$ is salt density; $\rho_r^0$ is core sample density before desalinization. Equation \ref{eq20:salt_conc} allows to estimate performance of ML models. In this case, data-driven algorithms should find hidden correlations within parameters and demonstrate appropriate predictive abilities. The appearance of the physical model gives an opportunity widely to test ML instruments.

For predictive models of porosity we took into account influence of all 10 characteristics of rock samples (Table \ref{tab:1}). Almost all surveyed methods (except Decision tree) has demonstrated promising results and high values of determination coefficient in the case of porosity prediction. Table \ref{tab:2} demonstrates results of models evaluation via cross-validation process: mean value and standard deviation. Here and further we reported linear regression only with L1 regularization, because of other implementations (without regularization and with L2 regularization) show very close results. However, this algorithm works slightly better. In general, the highest value of R2-metric corresponds to the lowest values of MSE and MAE metrics. In porosity case SVM, Neural network, Gradient boosting and Linear regressions have the best scores. The best two models are Support Vector Machines with $\textrm{R2} = 0.86 \pm 0.14, \: \textrm{MAE} = 0.82 \pm 0.19, \: \textrm{MSE} = 1.63 \pm 1.47$ and Neural network with $\textrm{R2} = 0.84 \pm 0.1, \: \textrm{MAE} = 0.94 \pm 0.16, \: \textrm{MSE} = 1.79 \pm 0.94$.

\begin{table}[ht!]
\caption{Results for porosity prediction}
\label{tab:2}       
\begin{tabular}{llllllll}
\hline\noalign{\smallskip}
No. & Model & $R2$ & $\sigma_{R2}$ & MAE & $\sigma_{MAE}$ & MSE & $\sigma_{MSE}$  \\ \hline
    1 & Linear regression with L1 regularization & 0.792 & 0.116 & 1.035 &  0.178 & 2.361 & 1.110\\ \hline
	2 & Decision tree & 0.490 & 0.226 & 1.749 & 0.322 & 5.882 & 2.244\\ \hline
    3 & Random forest & 0.675 & 0.090 & 1.427 & 0.213 & 3.815 & 1.073\\ \hline
    4 & Gradient boosting & 0.763 & 0.078 & 1.173 & 0.192 & 2.791 & 0.943\\ \hline
    5 & Gradient boosting (XGBoost) & 0.782 & 0.081 & 1.112 & 0.186 & 2.526 & 0.840\\ \hline
    6 & Support Vector Machines & 0.855 & 0.144 & 0.816 & 0.194 & 1.634 &  1.472\\ \hline
    7 & Neural Network & 0.841 & 0.098 & 0.935 & 0.164 & 1.793 & 0.935\\
\noalign{\smallskip}\hline
\end{tabular}
\end{table}

Grid search calculated optimal regularization value of L1 linear regression which equals 0.001. In a similar search process, optimal depth of decision tree was obtained - 7 and the optimal number of estimators (trees) for random forest - 150 along with the maximum depth of each tree - 8.  For Gradient Boosting model the following parameters were selected: 16000 estimators (trees), maximum depth of each tree - 2, subsample - 0.7 (it means that each tree takes only 70\% of initial data to fit, the next tree takes another 70\% randomly etc., this idea helps to prevent overfitting), max-features - 0.9 (the concept is similar to subsample, the only difference is that instead of using part of the samples, algorithm takes part of features to fit each tree), regularization - 0.001. Neural Network architecture was also defined with the help of grid search since we have the small dataset and can calculate several architectures fast. It has 2 hidden layers with  2 and 4 nodes in each layer. SVM was built with Gaussian kernel which has two parameters to tune: C and gamma. The exhaustive search showed optimal gamma - 0.0001 and optimal C - 40000. 
 
\begin{figure}[ht!]
\begin{center}
\begin{minipage}{.4\linewidth}
\subfloat[]{\label{main:a}\includegraphics[scale=.28]{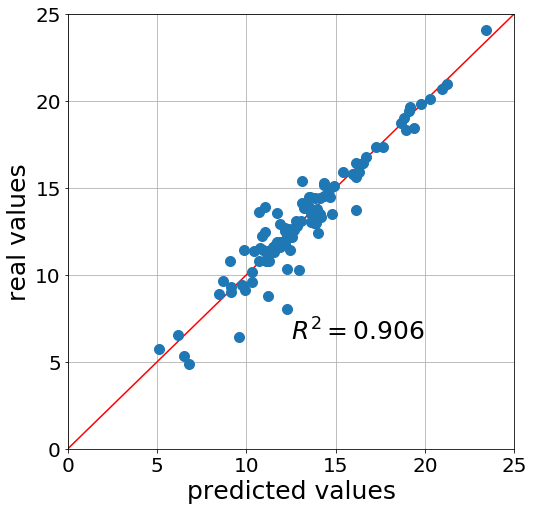}}
\end{minipage}%
\begin{minipage}{.4\linewidth}
\subfloat[]{\label{main:b}\includegraphics[scale=.28]{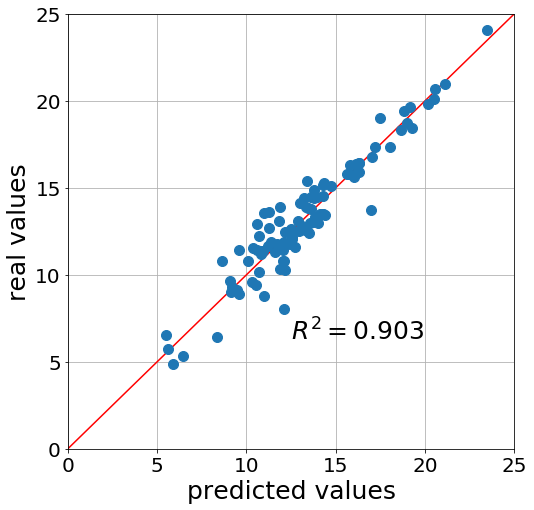}}
\end{minipage}\par\medskip
\caption{Comparison of actual and predicted porosity. (a) – for SVR algorithm, (b) – for Neural Network algorithm}
\label{fig3}
\end{center}
\end{figure}

Performance of the SVR and MLPRegressor models could be demonstrated by plotting predicted values (via cross-validation) of porosity after ablation versus the actual values (figure ~\ref{fig3}). One can see that the data points located along the mean line (bisectrix).

\begin{figure}[h]
\includegraphics[width=0.8\textwidth]{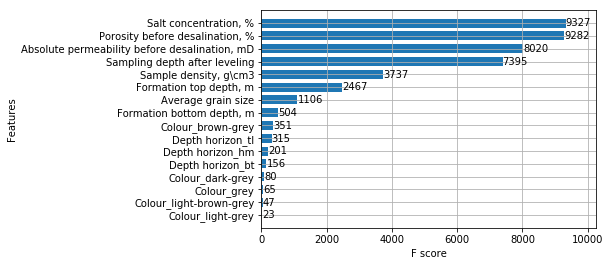}
 \caption{Feature Selection in porosity model with XGBoost}
 \label{fig4}
\end{figure}

Python’s XGBoost allows arranging features concerning  a degree of their influence on the prediction model \citep{Friedman2000Greedy}, see figure ~\ref{fig4}. The importance provides a score (referred as F score) that indicates how useful each feature was in the construction of the boosted decision trees within the model. This metric shows how many times the feature was used to split tree on \citep{freedman2009statistical}. The feature importance is then averaged across all of the decision trees within the model.

In figure ~\ref{fig4} one can see that porosity and permeability before desalination and salts concentration have the most significant influence on the porosity prediction results. What is in a good correspondence with the geometric correlation between porosity and salts concentration (equation \ref{eq20:salt_conc}). Core sample density before desalinization ($\rho_r^0$) also among the five influential features in ML algorithms (Figure \ref{fig4}). These observations demonstrate that predictive ML models simulate the same correlations as the physical model. A strong influence of the permeability before desalinization ($K_0$) on the porosity after desalinization ($\phi$) can be explained by a strong correlation between $K_0$ and $\phi_0$ (as, for example, in Kozeny-Carman equation form \citep{carman1956flow}). The next few features on Figure \ref{fig4}, which also significantly influence on porosity increase, are depths of sample, formation top and formation bottom. This fact confirms that the salts are distributed in formation non-uniformly and the distribution strongly depends on the geological condition of the reservoir. The colour features and the horizon types have the lowest influence on the prediction models.

\subsection{Permeability prediction}
\label{sec:3.2}
For permeability prediction all the methods also look promising and demonstrate high values of R2-metrics (table ~\ref{tab:3}). In this case Support Vector Machines, Neural network and linear regressions show the best performance. The highest scores were obtained for SVR with $\textrm{R2} = 0.86 \pm 0.08, \: \textrm{MAE} = 105 \pm 20, \: \textrm{MSE} = 39957 \pm 17906$ and Linear regression with $\textrm{R2} = 0.85 \pm 0.07, \: \textrm{MAE} = 118 \pm 20, \: \textrm{MSE} = 40864 \pm 16434$

\begin{table}[ht!]
	\centering
    \caption{Results for permeability prediction}
    \label{tab:3}
\begin{tabular}{llllllll}
\hline\noalign{\smallskip}
    No. & Model & R2 & $\sigma_{R2}$ & MAE & $\sigma_{MAE}$ & MSE & $\sigma_{MSE}$  \\ \hline
    1 & Linear regression with L1 regularization & 0.852 & 0.074 & 118 &  20 & 40864 & 16434\\ \hline
	2 & Decision tree & 0.677 & 0.196 & 167 & 42 & 90988 & 47737\\ \hline
    3 & Random forest & 0.775 & 0.103 & 139 & 36 & 68091 & 40814 \\ \hline
    4 & Gradient boosting & 0.806 & 0.085 & 139 & 33 & 57696 & 29888\\ \hline
    5 & Gradient boosting (XGBoost) & 0.809 & 0.093 & 137 & 36 & 57345 & 36071\\ \hline
    6 & Support Vector Machines & 0.856 & 0.078 & 105 & 20 & 39957 & 17906\\ \hline
    7 & Neural Network & 0.850 & 0.123 & 108 & 34 & 40256 & 32030\\ 
    \noalign{\smallskip}\hline
    \end{tabular}
\end{table}

Similar to the previous section, one may define optimal hyperparameters of algorithms for permeability prediction via grid search \citep{bergstra2012random}. 
So, for Linear regression model, we have used L1 regularization with regularization parameter equals to 10. 
The decision tree was built with maximum depth - 10.
25 trees with the maximum depth of 8 were defined for Random Forest algorithm.
In Gradient boosting model we obtained the following optimal values: 300 estimators with a maximum depth of each tree equals 2. Only 80\% of the samples and 90\% of the features have been used for each tree to fit the model and regularization parameter - 0.1. 
Neural network had 2 hidden layers with 77 and 102 nodes in each layer. 
SVM parameters included the Gaussian kernel with gamma equals to 0.0001 and C equals to 50000.

\begin{figure}[h]
\begin{center}
\begin{minipage}{.4\linewidth}
\subfloat[]{\label{main:a}\includegraphics[scale=.28]{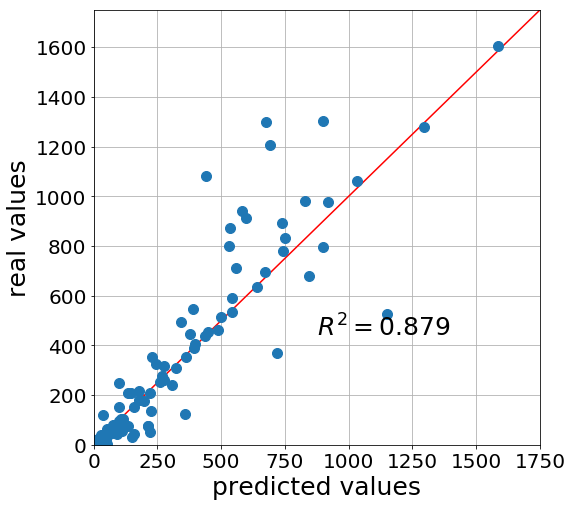}}
\end{minipage}%
\begin{minipage}{.4\linewidth}
\subfloat[]{\label{main:b}\includegraphics[scale=.28]{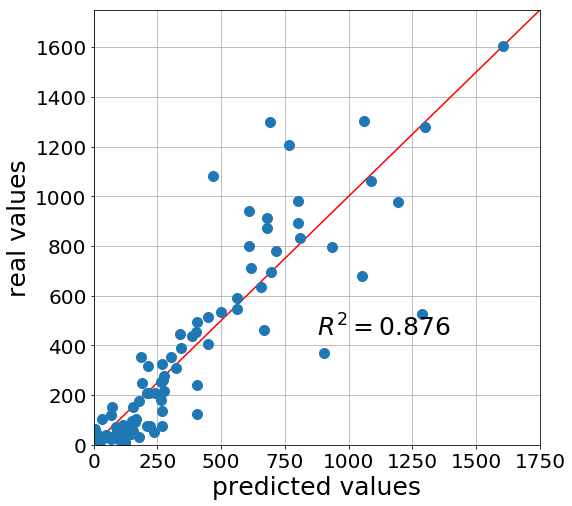}}
\end{minipage}\par\medskip
\caption{Comparison of real and predicted permeability. (a) – for SVR algorithm, (b) – for Linear Regression algorithm}
\label{fig5}
\end{center}
\end{figure}

Performance of the SVR and Linear regression models is demonstrated on the mean line plots in figure \ref{fig5}. Similarly, the prediction was made via cross-validation for all points. Data points are mainly located along the bisectrix, but generally matching between observed and predicted permeability is weaker than in porosity case.

\begin{figure}[h]
\includegraphics[width=0.8\textwidth]{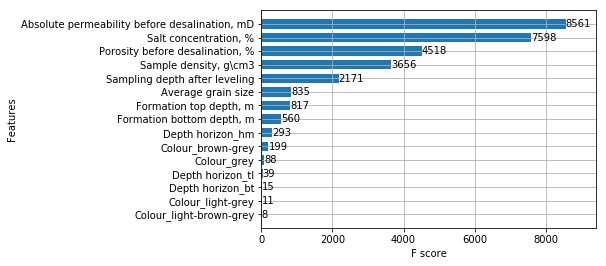}
 \caption{Feature Selection in permeability model with XGBoost}
 \label{fig6}
\end{figure}

The XGBoost method was also used to arrange features concerning their influence on the predictive model. In figure ~\ref{fig6} one can see that porosity and permeability before desalination and salts concentration have the most influence on the permeability prediction results. It is very similar to the results of Feature Selection in porosity model. We also obtained that the next features, which significantly influence permeability increase, are connected with the geological condition of the reservoir (sample depth, formation top and bottom depths). The colour features and the horizon type also occurred in the lowest influencers.

\subsection{Salts concentration prediction}
\label{sec:3.3}
The last part of the research is devoted to the prediction of salts concentration. The models work worse and demonstrate rather weak performance with R2-metric hardly reaching 0.6. Only a few methods look promising and demonstrate reasonable values of R2-metrics (Table ~\ref{tab4}). The best algorithms are Neural network, Gradient boosting and Random forest. Linear regression and Decision tree models are unacceptable with very small R2-metrics. R2 for Support Vector Regression reached almost 0.5.  The best two models with the highest scores are Neural network (from MLPRegressor model of Scikit-learn) with $\textrm{R2} =0.66 \pm 0.25, \: \textrm{MAE} =0.77 \pm 0.18, \: \textrm{MSE} = 1.69 \pm 0.96$ and Gradient boosting with $\textrm{R2} =0.59 \pm 0.27, \: \textrm{MAE} =0.93 \pm 0.19, \: \textrm{MSE} = 2.23 \pm 1.13$. Also, in this case, very high standard deviation (up to 100\%) in defining of R2, MSE and MAE metrics are obtained. This could be explained by non-uniformity of the experimental data.

\begin{table}[ht!]
	\centering
    \caption{Results for salts concentration prediction}
    \label{tab4}
\begin{tabular}{llllllll}
\hline\noalign{\smallskip}
    No. & Model & R2 & $\sigma_{R2}$ & MAE & $\sigma_{MAE}$ & MSE & $\sigma_{MSE}$  \\ \hline
    1 & Linear regression with L1 regularization & 0.014 & 0.385 & 1.579 & 0.174 & 5.747 & 1.996\\ \hline
	2 & Decision tree & 0.253 & 0.566 & 1.092 & 0.279 & 4.084 & 2.602\\ \hline
    3 & Random forest & 0.528 & 0.287 & 1.021 & 0.177 & 2.610 & 1.213\\ \hline
    4 & Gradient boosting & 0.593 & 0.265 & 0.929 & 0.191 & 2.227 & 1.131 \\ \hline
    5 & Gradient boosting (XGBoost) & 0.565 & 0.299 & 0.950 & 0.189 & 2.276 & 1.275\\ \hline
    6 & Support Vector Machines & 0.484 & 0.411 & 0.951 & 0.1909 & 2.608 & 1.221\\ \hline
    7 & Neural Network & 0.664 & 0.251 & 0.774 & 0.175 & 1.686 & 0.959\\ 
    \noalign{\smallskip}\hline
    \end{tabular}
\end{table}

By analogy with porosity and permeability, we defined optimal hyperparameters of algorithms via grid search process \citep{bergstra2012random}. Regularization parameter for L1 regression is equal to 1.0. The optimal decision tree has a depth of 9. Random Forest performs better with 10 estimators of the depth of 1. Gradient Boosting model runs with 300 estimators of the depth of 10. 95\% of samples and 50\% of features have been used for training of each tree. Regularization parameter is small and equals to 0.00001. The neural network contains 3 layers and 55, 10, 86 nodes in each layer respectively. SVM was performed with gamma equals to 0.1 and C equals to 25.

\begin{figure}[ht!]
\begin{center}
\begin{minipage}{.4\linewidth}
\subfloat[]{\label{main:a}\includegraphics[scale=.28]{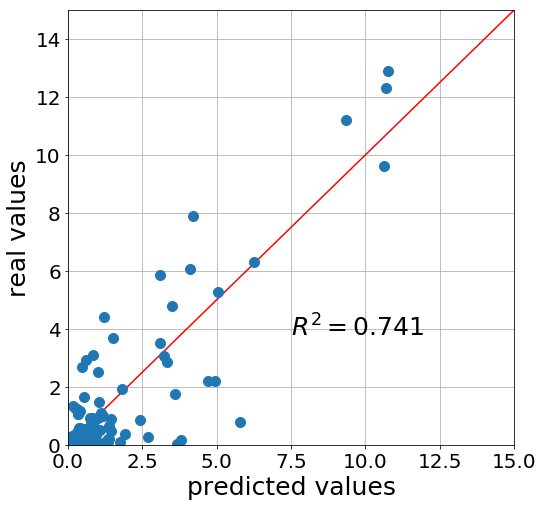}}
\end{minipage}%
\begin{minipage}{.4\linewidth}
\subfloat[]{\label{main:b}\includegraphics[scale=.28]{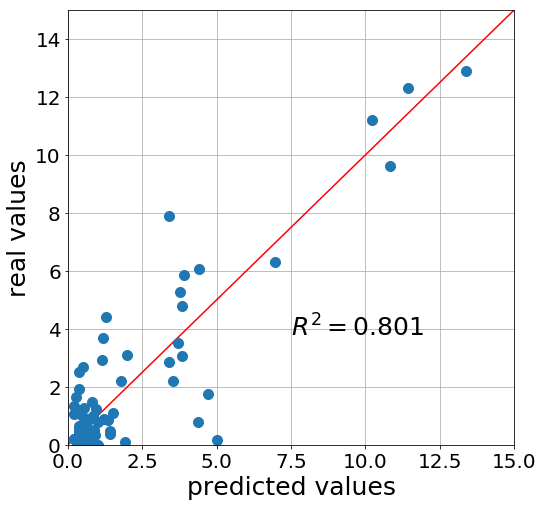}}
\end{minipage}\par\medskip
\caption{Comparison of real and predicted salts concentration. (a) – for Gradient Boosting algorithm, (b) – for Neural Network algorithm}
\label{fig7}
\end{center}
\end{figure}

Performance of the Gradient boosting and Neural network models are demonstrated on the mean line plots (figure ~\ref{fig7}). Data points partially located along the mean line. Accordingly, the correlation between observed and predicted values is much weaker than in porosity and permeability cases.

\begin{figure}[ht!]
\includegraphics[width=0.8\textwidth]{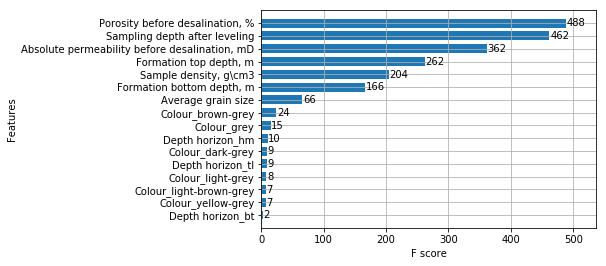}
 \caption{Feature Selection in salts concentration model with XGBoost}
 \label{fig8}
\end{figure}

Results of XGBoost features arrangement is in figure ~\ref{fig8}. As one can see, porosity before desalination has the most substantial influence on the salts concentration prediction results. The next two features affecting the prediction results are sample depth and permeability before desalination. Results of this Feature Rating differs from the results obtained for porosity and permeability models. We can state that the prediction of porosity and permeability alteration is primarily controlled by its initial values and amount of salts in the pore volume. Salts concentration, in its turn, strongly depends not only on the initial porosity and permeability but also on the formation pattern characteristics, which are linked with post-sedimentation processes. Therefore, the prediction model attempts to learn from training dataset where and how strong these processes are developed in the certain reservoir beds with various depth and location in the oilfield (through the formation top and bottom depths).

\subsection{Comparison of the predictive models with traditional approaches}
\label{sec:3.4}
\begin{figure}[h!]
\centering
\includegraphics[width=1\textwidth]{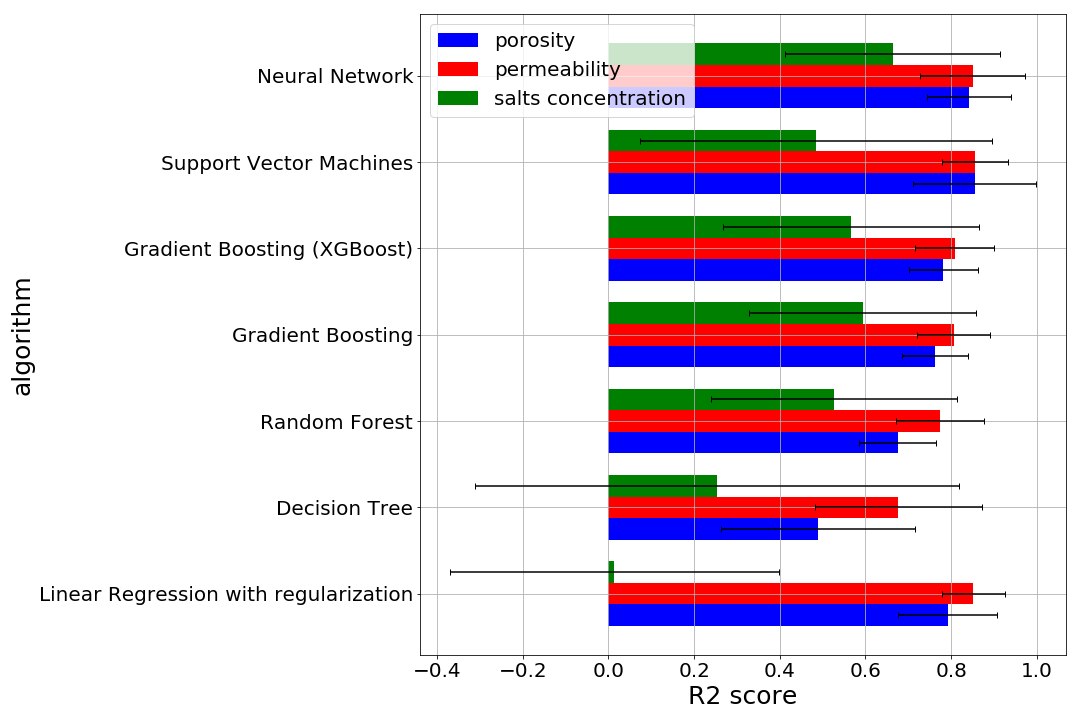}
 \caption{R2 scores for all models}
 \label{fig9}
\end{figure}

All obtained R2 scores with its variances for all algorithms are represented in figure  ~\ref{fig9}. The worst results could be associated with Decision tree method where we obtained not only the lowest values for R2 metric but the largest standard deviation of R2. Support Vector Machines and Linear regression demonstrate good results only for porosity and permeability prediction, but these methods are inappropriate for salts concentration prediction. The best machine learning method for prediction of all three petrophysical characteristics is Neural network in MLPRegression implementation. This algorithm demonstrates the most significant values of R2 metrics and the smallest standard deviation. Gradient boosting and Random forest could also be recommended as effective methods for prediction of salts concentration and permeability and porosity alteration due to salts ablation.

\begin{figure}[h!]
\centering
\includegraphics[width=1.0\textwidth]{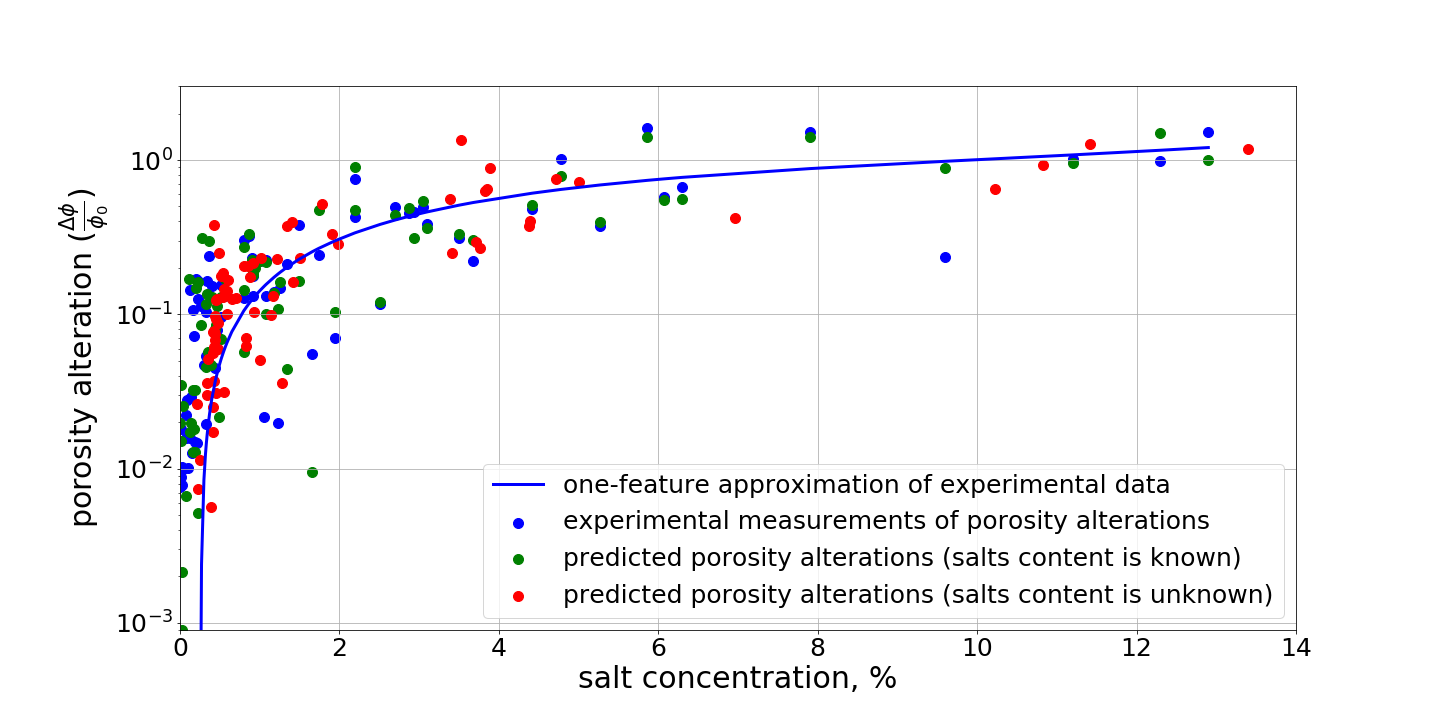}
 \centering
 \caption{Dependence of real and predicted porosity alteration on salts content. $\Delta\phi = \phi - \phi_0$, where $\phi_0$ is porosity before salts ablation and $\phi$ is porosity after salts ablation.}
 \label{fig10}
\end{figure}

\begin{figure}[h!]
\centering
\includegraphics[width=1.0\textwidth]{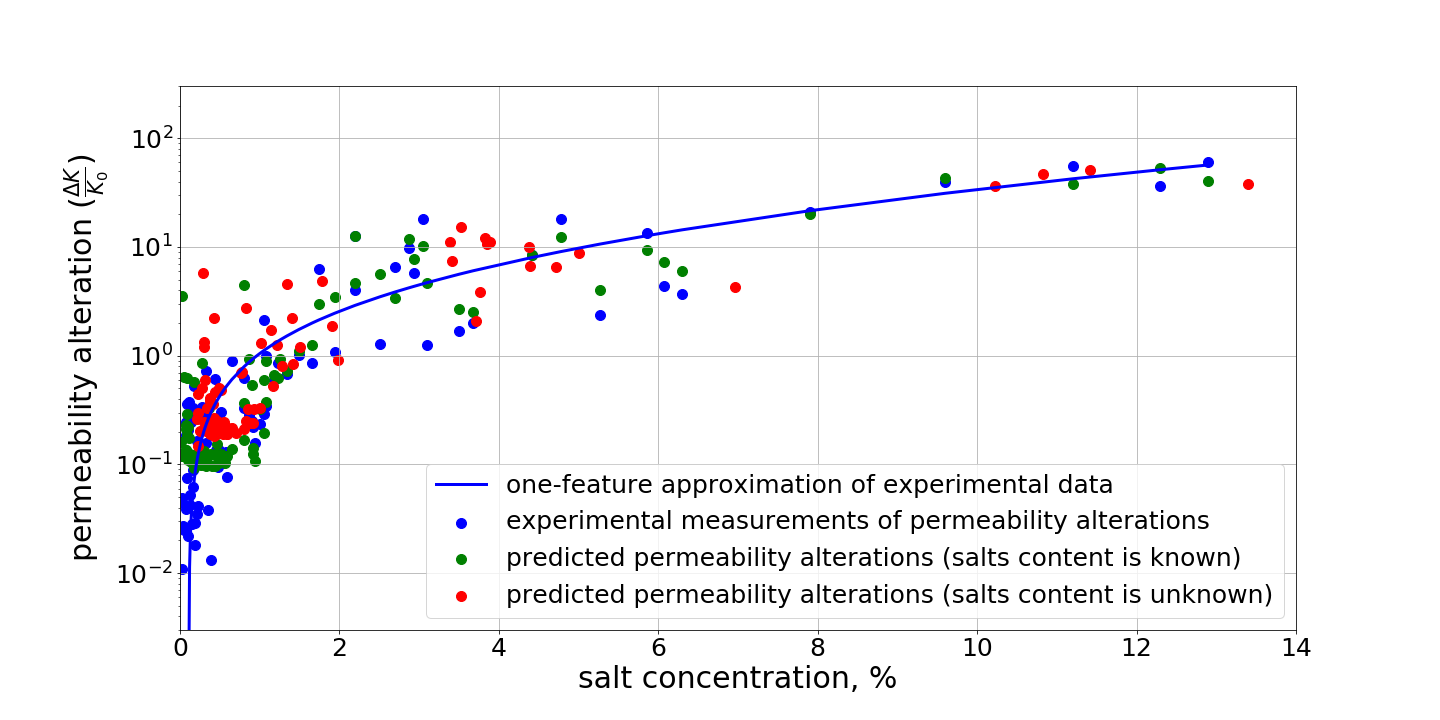}
 \centering
 \caption{Dependence of real and predicted permeability alteration on salts content. $\Delta K = K - K_0$, where $K_0$ is permeability before salts ablation and $K$ is permeability after salts ablation.}
 \label{fig11}
\end{figure}

The benefits of using machine learning models to estimate rock properties in comparison with standard one-feature approximation are obvious. When we talk about standard one-feature approximation, we assume the next approach. In case we do not know the law (or physical model) controlling the correlation between core sample characteristics and the target rock property the simplest and the fastest way to build the petrophysical model of the property is consistent single variable function analysis - finding consistently the target rock property functional dependence on each variable (characteristic of rock). The best single variable correlation in this approach could be considered as a one-feature approximation model. Instead of this expert approach ML algorithms allow building multi-feature approximations, which are more relevant to real rock properties correlations. Using one-feature approximation analysis, we found that porosity and permeability after ablation have the strongest correlation with salts concentration and corresponding dependencies showed in figures ~\ref{fig10} and ~\ref{fig11}. However, it does not mean that other core sample characteristics are useless. Over against, ML algorithms accounting all 10 characteristics should demonstrate better results.
In our case, the one-feature approximation is the cubic polynomial which accounts for the dependency of porosity (permeability) alteration on salt content.

To compare ML methods with one-feature approximation approach predictions of porosity and permeability alterations were performed in three ways:
\begin{itemize}
\item using one-feature approximations for porosity and permeability dependencies on salts content (here we take into account only one feature, solid lines in figures ~\ref{fig10} and ~\ref{fig11}); 
\item using prediction models based on Neural network with all 10 predetermined features (dataset with salts concentration measurements, green dots in figures ~\ref{fig10} and ~\ref{fig11});
\item using prediction models based on Neural network with only 9 predetermined features (dataset without salts concentration measurements, red dots in figures ~\ref{fig10} and ~\ref{fig11}).
\end{itemize}

The last approach includes a two-step procedure. First, we estimate salts concentrations with corresponding prediction model and second, use this predicted values in porosity and permeability predictions. This approach is applicable in the case when we do not have experimental measurements of salts concentrations.

\begin{figure}[h!]
\centering
\includegraphics[width=0.7\textwidth]{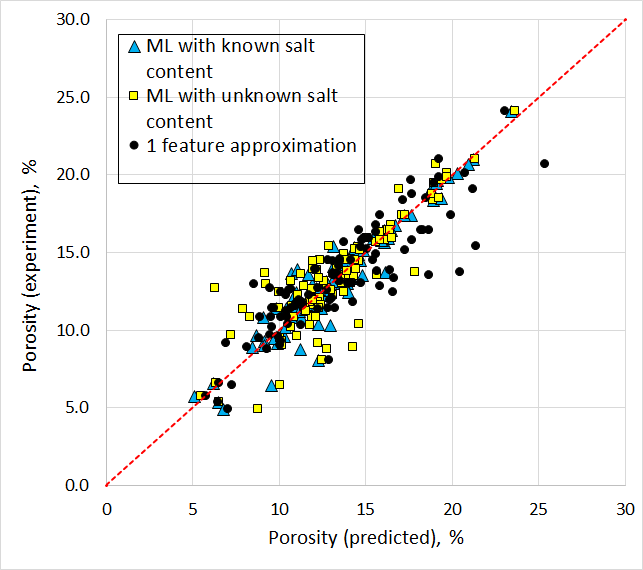}
 \centering
 \caption{Comparison of prediction by ML model and one-feature approximation for porosity}
 \label{fig12}
\end{figure}

\begin{figure}[h!]
\centering
\includegraphics[width=0.7\textwidth]{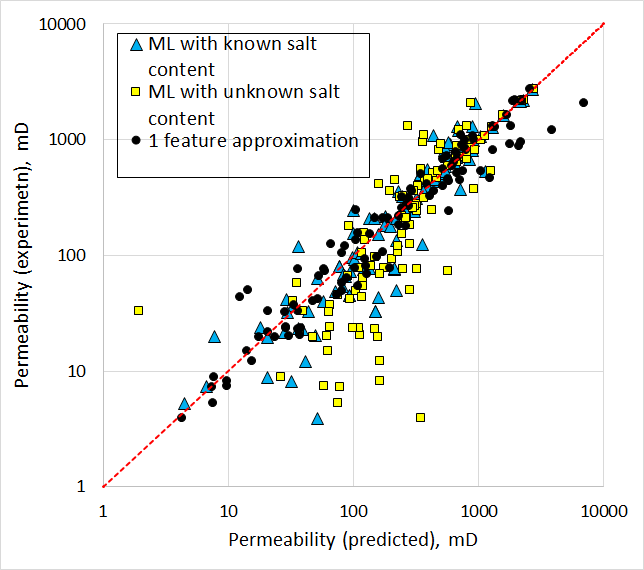}
 \centering
 \caption{Comparison of prediction by ML model and one-feature approximation for permeability}
 \label{fig13}
\end{figure}

The more detailed comparison of machine learning models and standard one-feature approximation presented in figures ~\ref{fig12}, ~\ref{fig13}. Here, the experimentally measured porosity (figure ~\ref{fig12}) and permeability (figure ~\ref{fig13}) were compared with the predicted values from ML (with and without salt content measurements) and one-feature approximation. These plots demonstrate the difference between three applied approaches for estimation of porosity and permeability after desalinization. Originally one feature approximation were obtained from dependency of porosity (permeability) alteration on salt content (figures ~\ref{fig10}, ~\ref{fig11}). Than these alterations were used to obtain values of porosity and permeability after desalinization. Blue triangles in figures ~\ref{fig12}, ~\ref{fig13} relate to ML model with known salt content. These points are located near the plot diagonal (ideal case). Black points relate to one-feature approximation, and these points are the most distant from diagonal. Yellow squares are approximation by ML with salt content preliminary predicted by ML. These predictions were made by using the approach described in section 2.4 in cross-validation as a third step (k-fold cross-validation with k equals to the number of samples). This method helps to compare approaches of permeability evaluation between each other and depict them on the plot. It does not evaluate the performance of ML algorithms well (methods evaluation was given in tables 2,3,4), because we have only one value of R2 (equation \ref{eq16:R2}) and do not have confidence intervals. From figures ~\ref{fig12}, ~\ref{fig13} one can see, that machine learning models work better.

Quantitative comparison of models is presented in Table \ref{tab5}. Negative value of R2 for the one-feature approximation of permeability was obtained because of several points, which after recalculation (from normalized permeability to absolute values) lay very far from experimental data and add huge error in R2 calculation (equation ~\ref{eq16:R2}). One may remove these points and obtain much better estimation, but ML models work with satisfying accuracy at these points, so, we've left results as it is to demonstrate the superiority of ML over old method.
Table \ref{tab5} confirm that simple polynomial regression taking into account only one feature at the same time works not so well as machine learning models considering many different features. We can also see that restriction of the dataset (case without salts concentration measurements) does not strongly affect prediction quality. However, it makes it possible to predict porosity and permeability alterations using only formation and core sample depths, initial porosity and permeability, rock density and lithology description. Feature ranking for salts concentration, permeability and porosity alterations models with Python's XGBoost method demonstrate that sample colour and horizon have a feeble influence on the predictive models and could be excluded from feature list for further applications. 

\begin{table}[ht!]
	\centering
    \caption{Performance of the prediction models}
    \label{tab5}
\begin{tabular}{lllll}
\hline\noalign{\smallskip}
    No. & Metric & ML (salts is known) & ML (salts is unknown) & one-feature approx.  \\ \hline
        & Porocity & & & \\ \hline
	1 & R2 & 0.90623 & 0.73156 & 0.69427 \\ \hline
    2 & MAE & 0.69695 & 1.20640 & 1.41261 \\ \hline
    3 & MSE & 1.1197 & 3.20539 & 3.65065 \\ \hline
        & Permeability & & & \\ \hline
    1 & R2 & 0.87924 & 0.80541 & -0.22504 \\ \hline
    2 & MAE & 95.923 & 140.877 & 188.569 \\ \hline
    3 & MSE & 37042.88 & 59687.87 & 375778.98 \\  \hline
    \end{tabular}
\end{table}

\section{Conclusion}
\label{sec:4}

In this paper applicability of various Machine Learning algorithms for prediction of some rock properties were tested. We demonstrated that three special properties of salted reservoirs of Chayandinskoye field could be predicted only basing on routine core analysis data. The target properties were:

\begin{itemize}
\item alteration of open porosity, 
\item alteration of absolute permeability,
\item salts mass concentration.
\end{itemize}

After core desalination permeability could be increased up to 60 times and porosity - up to 2.5 times. Usually these characteristics are out of RCA scope because it is time-consuming and occasional analysis. It is very useful for reservoir development planning to have the predictive models in case of lack of this type of data. Porosity and permeability before desalination, sample density, lithology and texture description are the RCA input data for our predictive models.

To build relevant predictive models the dataset with results of 100+ laboratory experiments was formed. The main 9 features were: formation top depth, formation bottom depth, initial (before desalination) porosity and permeability, sample depth adjusted to log depth, sample density (before desalination), average grain size (by lithology and texture description), sample colour and horizon ID. These features were used to build the salts concentration predictive model. For porosity and permeability alteration prediction we additionally used the 10th feature – the salts concentration. From a technical point of view, there is no matter these concentrations measured or predicted with other ML model.
    We reported 7 algorithms:
\begin{itemize}
\item Linear regression with L1 regularization;
\item Decision tree;
\item Random Forest;
\item Gradient boosting;
\item XGBoost;
\item Support vector machine;
\item Artificial neural network.
\end{itemize}    
    
The best two algorithms for porosity and permeability alteration prediction were Support Vector Machines with and Neural network. For permeability the Linear regression with regularization also showed good results. The best models demonstrate the determination coefficient R2 of 0.85+ for porosity and permeability. High precision of developed models looks to be helpful in decreasing of geological uncertainties in modelling of salted reservoirs. It was shown, that porosity and permeability before water intrusion along with the matrix density, sample depth and salts content are the most influencing features on permeability and porosity alteration.

The predictive model of salts concentration has been developed using the results of routine core analysis and data on core depth and top and bottom depths of productive horizons. The best algorithms here were Gradient boosting and Neural network. The highest coefficient of determination R2 for salts concentration in rocks equals 0.66. The precision of salts model is lower than the precision of porosity and permeability models. Nevertheless, the developed models allows to estimate the salts content in rocks without special experiments.

Combining all three models, it is also possible to make precise porosity and permeability alterations predictions using only a minimal volume of routine core analysis data: formation and core sample depths, initial porosity and permeability, rock density and lithology description. Accordingly, with these instruments geocientists and reservoir engineers can estimate the porosity and permeability alteration at waterflooding conditions having RCA measurements only.

It was shown that different algorithms work better in different models. However, the best machine learning method for prediction of all three parameters was two hidden layer Neural network in MLPRegression implementation. This algorithm gave the highest values of R2 metric and the smallest standard deviation. Gradient boosting and Random forest could also be recommended as alternative methods for predictions but with lower precision.
    
Finally, this work showed that machine learning methods could be applied for the prediction of rock properties, which laboratory measurements are time-consuming and expensive.


\bibliographystyle{spbasic}      
\bibliography{sourses.bib}   

%
%

\end{document}